\documentclass[letterpaper, 10 pt, conference]{ieeeconf}  % Comment this line out if you need a4paper

\IEEEoverridecommandlockouts                              % This command is only needed if 
\usepackage{graphics}
\usepackage{graphicx} % for pdf, bitmapped graphics files
\usepackage{amsmath} % assumes amsmath package installed
\usepackage{amssymb}  % assumes amsmath package installed
\usepackage{caption}
\usepackage{lettrine}

\title{\LARGE \bf
A Novel Feature Selection and Extraction Technique for Classification}

\author{Kratarth Goel$^{1}$, Raunaq Vohra$^{2}$ and Ainesh Bakshi$^{3}$
\thanks{$^{1}$Kratarth Goel is with the Department of Computer Science, BITS Pilani KK Birla Goa Campus, Goa, India
        {\tt\small kratarthgoel@gmail.com}}%
\thanks{$^{2}$Raunaq Vohra is with the Department of Mathematics, BITS Pilani KK Birla Goa Campus, Goa, India
        {\tt\small ronvohra@gmail.com}}%
\thanks{$^{3}$Ainesh Bakshi is with the Department of Computer Science, Rutgers University, New Brunswick, NJ
		{\tt\small aineshbakshi@gmail.com}}%
}

\begin{document}

\maketitle
\thispagestyle{empty}
\pagestyle{empty}

%%%%%%%%%%%%%%%%%%%%%%%%%%%%%%%%%%%%%%%%%%%%%%%%%%%%%%%%%%%%%%%%%%%%%%%%%%%%%%%%
\begin{abstract}
This paper presents a versatile technique for the purpose of feature selection and extraction - Class Dependent Features (CDFs). We use CDFs to improve the accuracy of classification and at the same time control computational expense by tackling the curse of
 dimensionality. In order to demonstrate the generality of this technique, it is applied to handwritten digit recognition and text categorization.
\end{abstract}

\smallskip
\noindent \textbf{\emph{Keywords-}} {\emph{MNIST; USPS; WebKB; Reuters-21578}}

% \fi
\IEEEpeerreviewmaketitle

\section{Introduction}

\lettrine{T}{his} paper proposes a novel and robust technique for feature selection and extraction which gives results comparable to the current state-of-the-art with the added advantage of being very fast and easy to implement on a range of devices. It suggests a better alternative to the Tf-Idf statistic, which can inadvertently decrease the statistic for words which occur frequently in a class label, and are innate to the entire class label, thereby resulting in poor features and lower accuracy for classification. The algorithm works by first selecting features relevant to their class label and extracts them accordingly. These extracted features are relevant to the entire class they are a part of, not just the individual data item they are extracted from. We call these features \emph{Class Dependent Features} (CDFs). Moreover the entire learning problem is then broken down into smaller classification tasks by creating a SVM for each pair of class labels. 
\section{The Algorithm}

\subsection{Feature Selection}

Consider a vector $P=\{P_1, P_2,.... ,P_m\}$ representing the set of all class labels in the training dataset ($m$ is the total number of class labels in the dataset). Each $P_c=\{p_1,p_2,....,p_M\}, p_k \in \mathbb{R\textsuperscript{\emph{N}}}, \forall k \in [1, M]$ further represents all data points in the $c\textsuperscript{th}$ class label. Then the function $f ({P_c}) = \{a_{c_1},a_{c_2},...,a_{c_N}\}$ representing the \emph{summation function} for the $c\textsuperscript{th}$ class label is given by

\begin{equation}
{a_{ci}} = \sum_{k=1}^{{M}}p_k(i)\>.
\end{equation}

where ${N}$ denotes the dimensionality of $p_k$,  ${M}$ is the cardinality of the $c\textsuperscript{th}$ class label and $a_{ci} \in \mathbb{R}$. Here, for instance, the element $p_k(i)$ would be the pixel intensity value of the $i\textsuperscript{th}$ pixel of the $k\textsuperscript{th}$ image, belonging to the class label $c$ (say, the digit $0$) of the MNIST dataset. We then obtain the measure $T ({P_c})=\{q_{c_1},q_{c_2},...q_{c_N}\}$ 

\begin{equation}
q_{ci} = a_{ci} / {M}\>
\end{equation}
$\>\>\>\>\>\> \forall i \in [1,N]$.

This measure \emph{T} represents the probability distribution over the entire digit class. 

Now that we have the measure \emph{T} for each class label, we must establish a relation $\mathbf{R_{xy}}$ between each pair of class labels $P_x$ and $P_y$. This is used to obtain the degree of relatedness between the two class labels $P_x$ and $P_y$, which gives us the degree of similarity and dissimilarity between them. The experiments in this paper were conducted taking $\mathbf{R_{xy}}$ as 

\begin {equation}
\mathbf{R_{xy}} = \{ q_{xi}/q_{yi}  \>\>| \>\>\forall q_{xi} \in T({P_x}) \>\> and \>\>\forall q_{yi} \in T ({P_y})  \}\> .
\end {equation}

We then take the mean of all the values of $q_{xi}/q_{yi} \in  \mathbf{R_{xy}}$ as shown below.
\begin {equation}
\mu_{xy} = \frac{\sum_{i=1}^{{N}}(q_{xi}/q_{yi})}{N}\> .
\end {equation}

We generate two thresholds $\tau$ and $\tau'$, given by

\begin{equation}
\begin {split}
\tau = b\mu_{xy} \> . \\
\tau' = b'\mu_{yx}\> .
\end {split}
\end{equation}

where $b,b' \in \mathbb{R}$ 

Values of $q_{ci} \in T(P_c)$ greater than the thresholds $\tau$ or $\tau'$ will be selected as the feature locations for the $c\textsuperscript{th}$ class label. Considering $T(P_c)$ as a vector of real values, only those indices \emph{i.e.} $i$'s in $T(P_c)$ for which  $q_{ci}$'s have their values greater than either threshold are chosen as class dependent. Hence the values $b$ and $b'$ can be thought of as parameters controlling dimensionality of the input space for the given problem statement.

Thus we define a new dataset $P'=\{P_1', P_2',.... ,P_m'\}$, where each $P_c'$ is the set of modified data items in the $c\textsuperscript{th}$ class label, given by 
\begin{equation}
\begin{split}
P_c'=\{p_1',p_2',....,p_M'\}\> .
\end{split}
\end{equation}

where each $p_k' , \forall k \in [1,M]$ is a probability distribution, formed as follows:

\begin{equation}
p_k'(i) = \left\{
        \begin{array}{ll}
            p_k(i) ,&if \> q_{ci} > \tau \>or \>q_{ci}  > \tau' \\
            NULL ,& \quad otherwise
        \end{array}
    \right.
\end{equation}
$\>\>\>\>\>\>\>\>\>\>\forall i \in [1,N]$.

Thus we have effectively reduced the dimensionality of the dataset by keeping only the class dependent values of these features \emph{i.e} the values of the features greater than the thresholds $\tau$ or $\tau'$. This can be interpreted as the non-NULL values represented in equ. (7). 

\subsection{Feature Extraction}

With our thresholds $\tau$ and $\tau'$ calculated, we can now proceed with the extraction of class dependent features (CDFs) for the pair of class labels $x$ and $y$. In order to do so, we use the concept of the \emph{Kullback-Leibler (KL) divergence}. The feature vector $\mathbf{F_{xy}}$ used for the purpose of classification between the pair of class labels $x$ and $y$ is calculated as follows:

\begin{equation}
\mathbf{F_{{xy}}}(k) = D_{KL}(p_k'\> ||\> T(P_x))\> .
\end{equation}

Also a set of labels is created to be given to the classifier as follows:

\begin{equation}
\mathbf{L_{{xy}}}(k)  = \left\{
        \begin{array}{ll}
             \>\>\>1 & \quad p_k' \in P_x'\> . \\
            -1 & \quad p_k' \in P_y'\> .
        \end{array}
    \right.
\end{equation}

We have now obtained our CDFs for each class label comparison. These features will be passed to the classifier for training on the dataset. This process is repeated for all pairs of class labels $x$ and $y$ to cover the entire dataset.

%------------------------------------------------------------------------
\section{Experimental Results}
%-------------------------------------------------------------------------

In order to test our technique and prove its generality and versatility, we used it on two fundamentally different problem statements - handwritten digit recognition and text categorization. For the former, we used the well-known MNIST and USPS datasets and used the WebKB and Reuters-21578 datasets for the latter. We used SVMs as the classifier for both problem statements and the optimum parameters were determined by using $n$-fold cross-validation.

\subsection{Handwritten Digit Recognition}

\begin{center}
	\begin{tabular}{|l|r|}

		\hline
		\bfseries{Techniques} & \bfseries{Error}\\
		\hline
		Linear classifier (1-layer NN) & 12.0\\
		K-nearest-neighbors, L3 & 2.83\\
		Products of boosted stumps & 1.26\\
		40 PCA + Quadratic classifier & 3.3\\
		SVM, Gaussian kernel & 1.4\\
		3-layer NN, 500+150 hidden units & 2.95\\
		2-layer NN, 800 HU, Cross-Entropy & 1.53\\
		Deep Belief Net & 1.0\\

Large Convolutional Net (no distortions) & 0.62\\
\hline
		\bfseries{CDFs, SVM, 2-degree poly kernel} & \bfseries{1.25}\\
		\hline
	
	\end{tabular}
	\vspace*{0.1cm}
\captionof{table}{Comparison of various generic techniques with error rates (\%) on the MNIST dataset.}

\end{center}

As can be observed from Table I - which shows the error rates for various technique on the MNIST dataset - our technique produces the best reported error rate among generic classification algorithms (no preprocessing was carried out on the data) after Hinton and Salakhutdinov's (2007) deep belief nets (1.00\%) and Deng and Yu's deep convex nets (0.83\%). It outperforms Kegl and Busa-Fekete's products of boosting stumps (1.26\%).

\subsection{Text Categorization}

The macro and micro averaged F measure on the Reuters-21578 dataset are tabulated in Table II. It is evident that CDFs outperform all other techniques and are a significant improvement upon them. The macro and micro F scores (89.28\% and 96.32\%) exceed Gini index, the second best, by 20\% and 6.5\% respectively.

These experiments were carried out on a 2\textsuperscript {nd} Generation Intel Core i5-2410M processor running Ubuntu 13.04 and the code was implemented using OpenCV 2.4.6.1 \cite{A20}. The training and testing on the MNIST dataset takes 5.77 minutes. In comparison, while running the MNIST task using Stacked Denoising Autoencoders, pre-training takes 585.01 minutes. Fine-tuning is completed after 36 epochs in 444.2 minutes. The final testing score obtained is 1.3\%, which our technique outperforms by 0.05\%. These results were obtained using Theano 0.6rc3 \cite{A35} on a machine with an Intel Xeon E5430 @ 2.66GHz CPU, with a single-threaded GotoBLAS. This serves to illustrate the accuracy our technique brings with its computation time being a fraction of that taken by Stacked Denoising Autoencoders.  

\begin{center}

	\begin{tabular}{|l|c|c|}

		\hline
		\bfseries{Feature Weight Function} & \bfseries{Macro F} & \bfseries{Micro F}\\
		\hline
		
		TFxIDF & 62.63 & 84.73\\
		TFxGINI & 69.82 & 89.79\\
		TFxIG & 63.88 & 84.64\\
		TFxCROSS-ENTROPY & 66.63 & 86.55\\
		TFxX2 & 60.88 & 83.71\\
		TFxMUTUAL-INFO & 68.15 & 87.59\\
		TFxODDS-RATIO & 69.16 & 88.05\\
		TFxWEIGHT OF EVID & 64.24 & 85.22\\
		\hline
		\bfseries{CDF} & \bfseries{89.29} & \bfseries{96.32}\\
		\hline
	\end{tabular}
	\vspace*{0.02cm}

\captionof{table}{Benchmarking CDFs with contemporary techniques on Reuters-21578 using SVM classifiers.}

\end{center}

%\addtolength{\textheight}{-11.8cm}

\section{Conclusions and Future Work}

We have proposed a technique for feature selection and extraction using \emph{Class Dependent Features} (CDFs). Our technique has been tested on the MNIST and USPS datasets for handwritten digit recognition as well as on the Reuters-21578 and WebKB datasets for text categorization. We have obtained strong competitive results using an SVM with a degree-2 polynomial kernel and these results along with those of contemporary techniques have been tabulated. Our results are comparable to the current state-of-the-art, and on par with the current best generic algorithms.

% Acknowledgments---Will not appear in anonymized version

{\small
\bibliographystyle{ieee}
\bibliography{egbib}

\end{document}